\documentclass[conference]{IEEEtran}
\usepackage{url}
\usepackage{cite}
\usepackage{indentfirst}
\IEEEoverridecommandlockouts
\usepackage{graphicx}
\usepackage{svg}
\usepackage{amsmath,amssymb,amsfonts}
\usepackage{algorithmic}
\usepackage{mathtools}
\usepackage{algorithm}
\usepackage{graphicx}
\usepackage{float}
\usepackage{textcomp}
\usepackage{subfigure} 
\usepackage{mwe}
\usepackage{textcomp}
\usepackage{float}
\usepackage{wrapfig}
\usepackage{rotating}
\usepackage{tikz}
\usepackage{textcomp}
\usepackage{mwe}
\usepackage{multicol}
\usepackage{multirow}
\usepackage{graphicx}
\usepackage{float}
\usepackage{amssymb}
\usepackage{gensymb}
\usepackage{array}
\usepackage{enumitem}
\usepackage{fancyhdr}
\usepackage{dirtytalk}
\usepackage{flexisym}
\usepackage{anyfontsize}
\usepackage[T1]{fontenc}
\usepackage{booktabs}
\usepackage{soul}

\usepackage[]{caption}

\usepackage{amsmath,amssymb,amsfonts}
\usepackage{algorithmic}
\usepackage{graphicx} 
\usepackage{textcomp}
\usepackage{xcolor}
\usepackage{rotating}
\usepackage{tikz}
\usepackage[hidelinks]{hyperref}
\usepackage{balance}
\usepackage{xspace}
\def\checkmark{\tikz\fill[scale=0.4](0,.35) -- (.25,0) -- (1,.7) -- (.25,.15) -- cycle;}

\newcommand{\linebreakand}{%
      \end{@IEEEauthorhalign}
      \hfill\mbox{}\par
      \mbox{}\hfill\begin{@IEEEauthorhalign}
    }

\newcommand{\ourmethod}{\textit{RadarTrack}\xspace}

\begin{document}
\title{\ourmethod: Enhancing Ego-Vehicle Speed Estimation with Single-chip mmWave Radar 
}

\author{\IEEEauthorblockN{Argha Sen\IEEEauthorrefmark{8}, Soham Chakraborty\IEEEauthorrefmark{2}, Soham Tripathy\IEEEauthorrefmark{3}, Sandip Chakraborty\IEEEauthorrefmark{4}}
\IEEEauthorblockA{Department of Computer Science and Engineering,
Indian Institute of Technology Kharagpur\\
Email: \IEEEauthorrefmark{8}arghasen10@gmail.com,
\IEEEauthorrefmark{2}sohamc1909@gmail.com,
\IEEEauthorrefmark{3}sohamtripathy2001@gmail.com,
\IEEEauthorrefmark{4}sandipchkraborty@gmail.com}\thanks{The first three authors have equal contributions.}
}

\maketitle

\begin{abstract}
In this work, we introduce \ourmethod{}, an innovative ego-speed estimation framework utilizing a single-chip millimeter-wave (mmWave) radar to deliver robust speed estimation for mobile platforms. Unlike previous methods that depend on cross-modal learning and computationally intensive Deep Neural Networks (DNNs), \ourmethod{} utilizes a novel phase-based speed estimation approach. This method effectively overcomes the limitations of conventional ego-speed estimation approaches which rely on doppler measurements and static surrondings. \ourmethod{} is designed for low-latency operation on embedded platforms, making it suitable for real-time applications where speed and efficiency are critical. Our key contributions include the introduction of a novel phase-based speed estimation technique solely based on signal processing and the implementation of a real-time prototype validated through extensive real-world evaluations. By providing a reliable and lightweight solution for ego-speed estimation, \ourmethod{} holds significant potential for a wide range of applications, including micro-robotics, augmented reality, and autonomous navigation.
\end{abstract}

\section{Introduction}
Comprehending the movements of mobile agents, whether navigating through distant planets or mixed reality environments, is fundamental for adequate perception and interaction. Ego-motion estimation is crucial, as it does not depend on pre-existing maps or environmental infrastructure. Instead, it derives the agent's position and orientation from the data collected during movement. Accurate odometry is also crucial for developing maps, especially when using techniques like \textit{Simultaneous Localization and Mapping} (SLAM) to dynamically build and update a model of the environment.


Due to their affordability and widespread availability, MEMS inertial sensors (IMUs) are widely utilized for ego-motion estimation across various mobile platforms. Estimating position from acceleration involves double integration, leading to errors~\cite{shen2018closing}. Accelerometers struggle to detect acceleration in ground vehicles moving at constant speeds due to subtle changes in acceleration~\cite{lee2020visual,yang2020online}. Moreover, IMU often requires precise calibration, which is challenging and labor-intensive. This calibration often requires specialized and costly equipment, such as synchronized clocks, precision turntables, and motion tracking systems. Multimodal odometry systems have been developed to address these limitations, integrating inertial data with additional sensor inputs, such as visual or ranging information. Among these, visual-inertial odometry (VIO) stands out for its robustness and is commonly found in devices like mobile phones, offering a more reliable solution for motion estimation~\cite{clark2017vinet,lee2020visual,qin2018vins}. However, VIO performance can suffer or even fail in difficult lighting conditions, such as in darkness where RGB cameras are ineffective or in intense illumination where depth cameras may encounter glare. Similar visibility challenges affect LiDAR-inertial odometry (LIO), especially in the presence of airborne particles such as dust, fog, or smoke. Moreover, LiDARs tend to be bulky, heavy, and costly, making them more suitable for high-end robotics than micro-robots or wearables.

mmWave based sensing offers distinct advantages over vision-based systems, particularly due to its robustness against environmental factors like scene illumination and airborne obscurants. Unlike mechanically scanning radars, single-chip mmWave radars utilize electronic beamforming, making them lightweight and ideal for micro-robots, mobile devices, and wearable technology. This technology is already being used for motion sensing in smartphones like Google Pixel~\cite{lien2016soli} or obstacle detection in commercial drones~\cite{yu2020autonomous,deng2022geryon}.


Traditional radar-based approaches estimate speed using doppler shifts, constrained by the Discrete Fourier Transform (DFT) to discrete steps of $\frac{\lambda}{2 N_c T_c}$, where $T_c$ is the chirp duration and $N_c$ is the number of chirps transmitted by the radar in a single frame~\cite{rao2017introduction}. Speeds not matching these steps incur errors, especially at very low speeds ($v < \frac{\lambda}{2 N_c T_c}$), placing them in the sub-doppler regime, as seen in microrobots.

\begin{figure}[!t]
    \centering
    \subfigure[]{
    \includegraphics[trim=25 15 25 25, clip,width=0.23\textwidth]{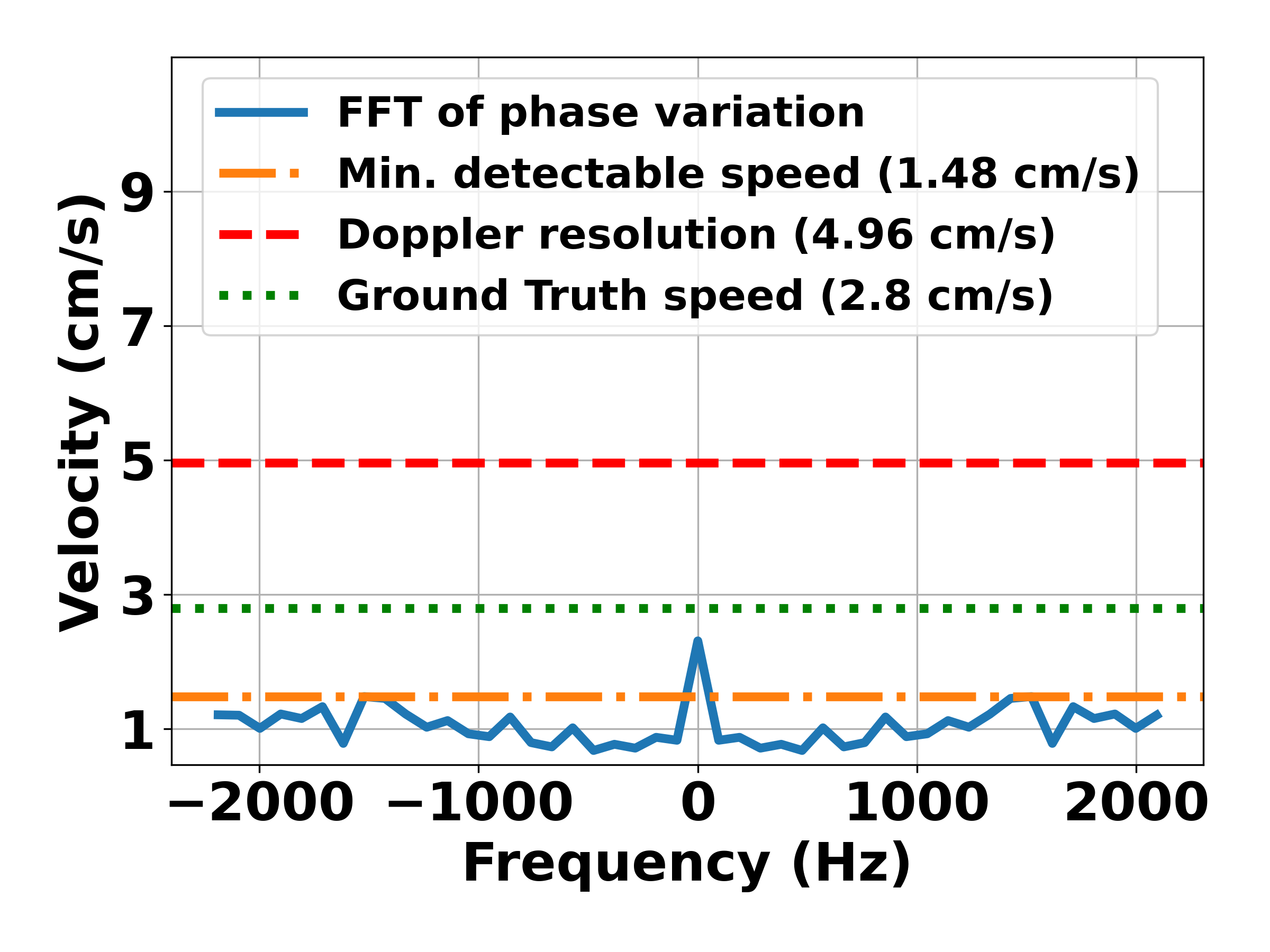}\label{fig:phasefft_slow}
    }\hfill
    \subfigure[]{
    \includegraphics[trim=25 15 25 25, clip,width=0.23\textwidth]{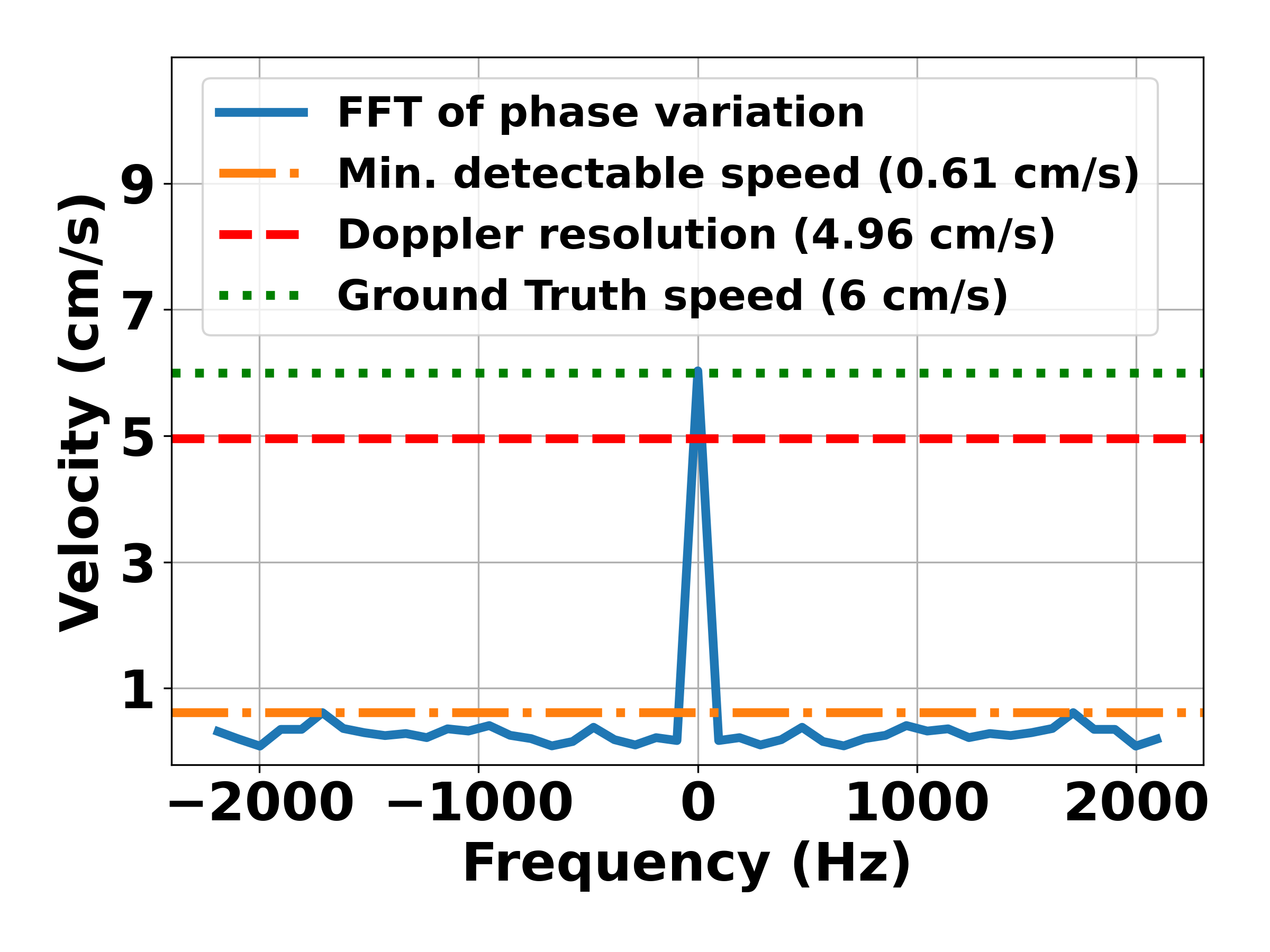}\label{fig:phasefft_high}
    }
    \caption{Comparison of doppler-based and phase-based speed estimation for a radar-mounted ego-vehicle moving at (a) 2.8 cm/s, and (b) 6 cm/s. The noise floor indicates the minimum achievable speed detection limit.}
    \label{fig:phase_motivation}
    \vspace{-0.5cm}
\end{figure}

To illustrate this limitation, we conducted a comparative study by mounting a radar on an ego-vehicle and placing a static object in front of it. By moving the ego-vehicle directly towards the static object, the radar captured phase variations from the reflected signals, and by applying FFT to these phase changes and scaling the results using the relation $v = \frac{\lambda}{4\pi} \frac{d\Phi}{dt}$, we obtained phase-based speed estimates~\cite{rao2017introduction}. \figurename~\ref{fig:phase_motivation} demonstrates the effectiveness of phase-based speed estimation compared to traditional doppler-based methods under different speed conditions. The blue line represents the FFT of phase variations from a static object in the radar's field of view (FoV), where the peak corresponds to the actual speed of the ego-vehicle. The noise level indicates the minimum achievable speed detection limit for the phase-based approach. 
As shown in \figurename~\ref{fig:phasefft_slow}, when the ego-vehicle operates in the sub-doppler regime, the doppler-based method will fail to resolve the motion accurately, whereas the phase-based approach successfully estimates the speed. In contrast, when the ego-vehicle speed exceeds the doppler resolution threshold, as shown in \figurename~\ref{fig:phasefft_high}, although the doppler-based method can estimate better, it still has resolution error, which is not the case for the phase-based approach, highlighting the advantage of phase-based speed estimation in scenarios where traditional doppler-based methods suffer from resolution constraints.

Creating a reliable indoor odometry system based on this concept faces several challenges. Firstly, radar returns are susceptible to noise from specular reflections, diffraction, and significant multipath effects. Also, due to hardware constraints on the number of antennas it results in highly sparse pointclouds (PCDs) with limited angular resolution. This low-quality data makes conventional LiDAR-based methods, such as \textit{Iterative Closest Point} (ICP)~\cite{lin2024icp}, almost ineffective when applied directly to mmWave radar data. 
Secondly, while some previous works~\cite{lu2020milliego,kramer2020radar,almalioglu2020milli, huang2024less, cen2018precise} have employed a multimodal approach by fusing radar data with other sensors, like IMUs and RGB cameras, the potential of mmWave radar to complement these modalities remains uncertain. Additionally, existing works perform well in environments with only static objects~\cite{sie2024radarize,kramer2020radar, cen2018precise} but fail to perform adequately in the presence of dynamic objects. Moreover, incorporating recent advances in DNNs for visual or LiDAR odometry poses challenges due to the significant computational load, which may limit their use in mobile, wearable, and other resource-constrained devices. Our research explores methods to make mmWave radar more computationally efficient, focusing on leveraging only the phase values of the reflected signals. In our previous work~\cite{sen2024poster}, we demonstrated that phase-based ego-speed estimation outperforms traditional approaches. However, this method is constrained by the requirement of a static environment and is effective only when a stationary object is directly near the boresight angle of the radar's FoV.  However, estimating the absolute ego-speed from phase variations of objects at oblique angles is not trivial. 

Motivated by these gaps of traditional ego-speed estimation, in this work, we propose \ourmethod{}, a novel ego-speed estimation approach relying solely on the phase returns of mmWave radar. We have developed a $4^{th}$-order expression of phase patterns in terms of the ego-speed and showcase how estimating the roots of this kinematics-based equation can estimate accurate speed. By leveraging a purely signal processing-based pipeline, \ourmethod{} enables us to achieve low latency in estimating ego-speed compared to its DNN counterparts. The contributions are summarized as follows:
\begin{enumerate}[leftmargin=*]
    \item \textbf{Kinematics-based analytic equation for phase variation:}
    We derive an equation to represent the phase change in objects induced by the ego-vehicle's motion as a fourth-order equation in terms of ego-speed, based on kinematics laws. By solving the roots of this equation we have precise ego-speed estimation even for static objects positioned at oblique angles in front of the radar.
    \item \textbf{Lightweight, radar-based ego-motion estimation across multiple platforms:}
    We present a novel ego-speed estimation framework implemented on three platforms- an Unmanned Ground Vehicle (UGV), a drone, and a handheld stick- using a single commercial-off-the-shelf (COTS) mmWave radar. Our method eliminates the need for traditional, bulky setups with multiple sensors.  
    \item \textbf{Dynamic and static object segmentation:} Our method introduces an innovative framework that combines doppler-based speed profiling to effectively segment dynamic and static objects in cluttered environments, helping us to robustly estimate ego-speed in highly dynamic environments. 
    \item \textbf{Real-time edge processing for efficient ego-speed estimation:} Our lightweight signal processing enables real-time performance without relying on deep learning models. With a latency of only $\approx 0.29$ s ($\approx 85\%$ lesser than its closest baseline, Radarize~\cite{sie2024radarize}) on a low-compute Jetson Nano, our method is ideal for mobile applications. 
\end{enumerate}


\section{Related Work}

We have classified the previous works on ego-motion estimation into two categories: (i) methods based on multi-modal fusion and (ii) methods that are purely radar-based.

\textbf{Multi-Modal Sensor Fusion:} Multi-modal sensor fusion combines data from various sensors, employing techniques from probabilistic fusion to deep learning-based attention mechanisms. Traditional methods like probabilistic fusion, fuzzy reasoning, and hybrid fusion have advanced with neural networks, allowing for data fusion at multiple levels—from raw data to final decisions \cite{almalioglu2020milli}. Data-driven deep odometry models have significantly advanced, allowing them to effectively model complex physical processes and disturbances through end-to-end learning. These models have been employed across various sensors, including RGB cameras~\cite{wang2017deepvo}, depth cameras~\cite{ummenhofer2017demon, yang2018deep}, IMUs~\cite{chen2018ionet,chen2019deep}, and LiDARs~\cite{lin2024icp}. MilliEgo~\cite{lu2020milliego}  proposes a trajectory estimation technique using either co-located multi-radar setups, radar-inertial data, or radar-inertial-visual data. It uses DNNs to regress out six degrees of freedom (DoF) pose estimates, incorporating a mixed attention multi-modal sensor fusion module to leverage the complementary nature of different sensors (mmWave, IMU, Camera, etc.). While it achieves promising results, it faces challenges with large drifts in handheld scenarios and imperfect loop closures. Additionally, the DL component leads to high inference times. The main advantage of deep odometry models is their ability to bypass extensive calibration, making them accessible to non-experts.

\begin{table}[]
\scriptsize
\centering
\caption{Comparison of SOTA approaches with \ourmethod. }\label{tab:rel_work}
\begin{tabular}{|l|l|l|l|l|l|}
\hline
SOTA Approaches       & Dynamic    & \begin{tabular}[c]{@{}l@{}}Radar\\ Only\end{tabular} & \begin{tabular}[c]{@{}l@{}}Doppler\\ based\end{tabular} & \begin{tabular}[c]{@{}l@{}}PCD\\ based\end{tabular} & \begin{tabular}[c]{@{}l@{}}Phase\\ based\end{tabular} \\ \hline
Milli-RIO~\cite{almalioglu2020milli}             & $\times$          & $\times$   (IMU)                                                  & $\times$                                                       & \checkmark                                                   & $\times$                                                     \\ \hline
Kramer et al~\cite{kramer2020radar} & $\times$          & $\times$  (IMU)                                                  & \checkmark                                                       & $\times$                                                   & $\times$                                                     \\ \hline
Radarize~\cite{sie2024radarize}              & $\times$          & \checkmark                                                    & \checkmark                                                       & $\times$                                                   & $\times$                                                     \\ \hline
EmoRI~\cite{ding2023push}                 & \checkmark          & \checkmark                                                    & \checkmark                                                       & \checkmark                                                   & $\times$                                                     \\ \hline
milliEgo~\cite{lu2020milliego}              & $\times$          & $\times$ (IMU)                                                   & $\times$                                                       & \checkmark                                                   & $\times$                                                     \\ \hline
Huang et al~\cite{huang2024less}  & \checkmark          & $\times$ (IMU)                                                    & \checkmark                                                       & \checkmark                                                   & $\times$                                                     \\ \hline
Cen et al~\cite{cen2018precise}    & $\times$          & \checkmark                                                    & $\times$                                                       & \checkmark                                                   & $\times$                                                     \\ \hline
4D iRIOM~\cite{zhuang20234d}              & \checkmark          & $\times$ (IMU)                                                   & $\times$                                                       & \checkmark                                                   & $\times$                                                     \\ \hline
radarTrack            & \textbf{\checkmark} & \textbf{\checkmark}                                           & $\times$                                                       & \checkmark                                                   & \textbf{\checkmark}                                            \\ \hline
\end{tabular}
\vspace{-0.5cm}
\end{table}


\textbf{Purely Radar-Based Methods:} Unlike these methods which rely on a combination of sensors, several recent works like~\cite{sie2024radarize,ding2023push,cen2018precise} focus solely on radar-based odometry using point association or doppler shift 
For example, Radarize~\cite{sie2024radarize} utilizes doppler-azimuthal heatmaps for translation estimation and range-azimuthal heatmaps for rotation estimation. The system processes consecutive heatmaps through neural networks to determine both the translation and rotation of the device, incorporating advanced techniques like echo suppression for multipath rejection. Despite achieving competitive accuracy, the method's reliance on neural networks and high computational requirements limit its real-time capabilities, especially in dynamic environments with numerous moving objects. In~\cite{sen2024poster} authors have proposed ego-speed estimation using raw phase values of the mmWave radar, however the proposed system can only work with a single static object and when the ego-vehicle is moving in a straight line towards the static object. Also, the estimated speed is the relative radial speed of the static object. \cite{ding2023push} focuses on improving ego-speed and trajectory estimation in environments with dynamic objects by employing a hybrid FFT-MUSIC algorithm to reduce angle estimation errors and selectively process data from static points to mitigate the influence of dynamic objects.~\cite{cen2018precise} incorporates advanced multipath rejection techniques, leveraging the shape consistency of PCDs across frames and determines the rigid body transformation that best aligns the PCDs. While effective in various scenarios, this approach requires careful handling of high-frequency noise and multipath reflections and may face limitations in environments with cluttered objects or significant measurement errors.

Also, works like~\cite{lu2020see,prabhakara2023high} deal with the inherent sparsity of the radar compared to optical sensors like LiDAR. 
\cite{lu2020see} employs GANs to densify sparse radar PCDs, while RadarHD~\cite{prabhakara2023high} uses an asymmetric UNet for radar heatmap upsampling. To the best of our knowledge, \ourmethod is the only dynamic, radar-only approach that utilizes PCD and phase-based features. In contrast, other SOTA methods rely on IMU, Doppler, or partial radar features (see \tablename~\ref{tab:rel_work}).
\begin{figure}[!t]
    \centering
    \includegraphics[width=0.5\textwidth]{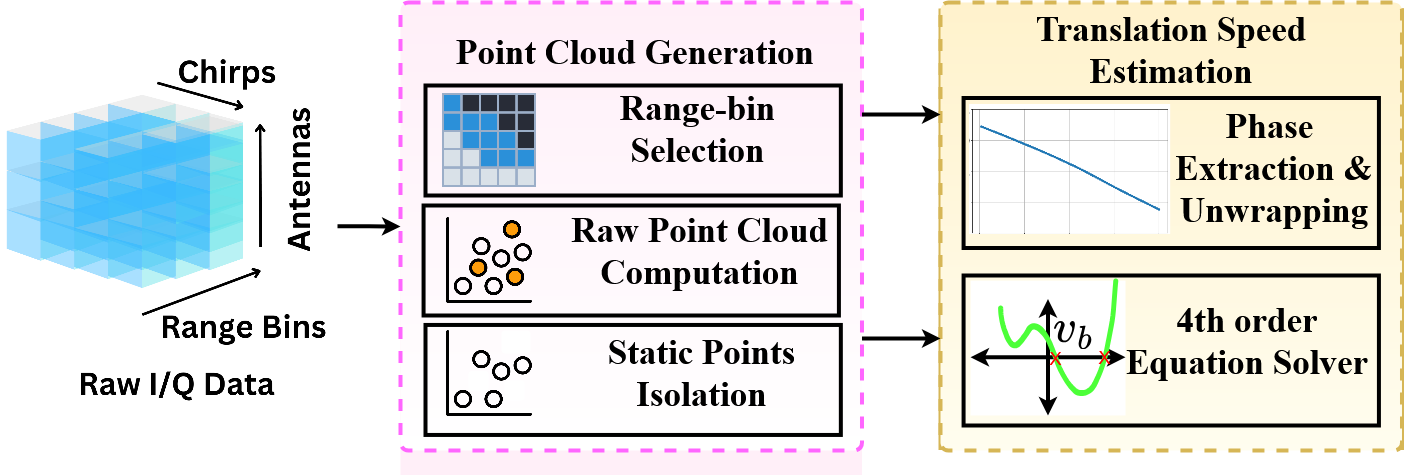}
    \caption{System pipeline of \ourmethod.}
    \label{fig:pipeline}
    \vspace{-0.5cm}
\end{figure}
\section{System Design}
Before we dive into the details of \ourmethod{}, let's outline the objectives and decisions that guided our design choices. 

\begin{figure*}[!t] 
    \centering
    \subfigure[]{
    \includegraphics[width=0.29\textwidth]{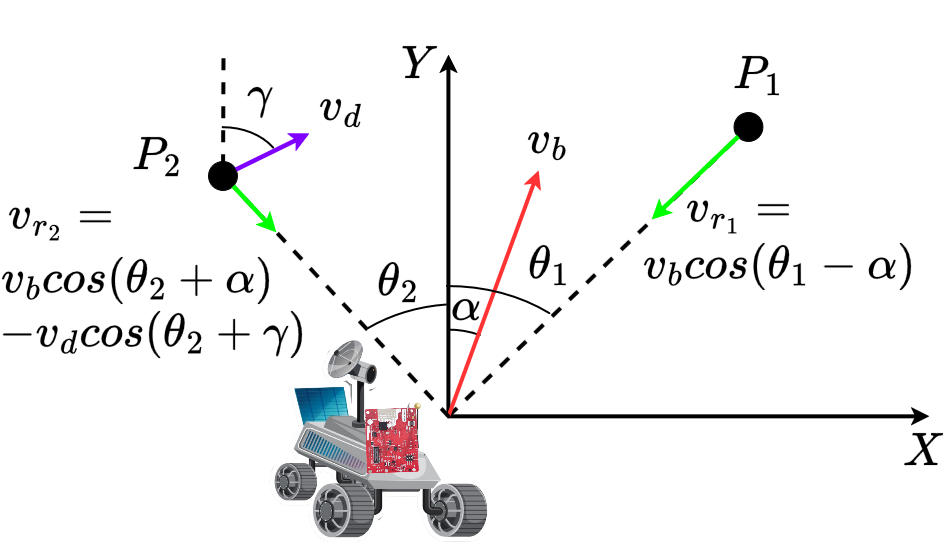}\label{fig:radial}
    }\hfill
    \subfigure[]{
    \includegraphics[trim=25 15 25 25, clip, width=0.22\textwidth]{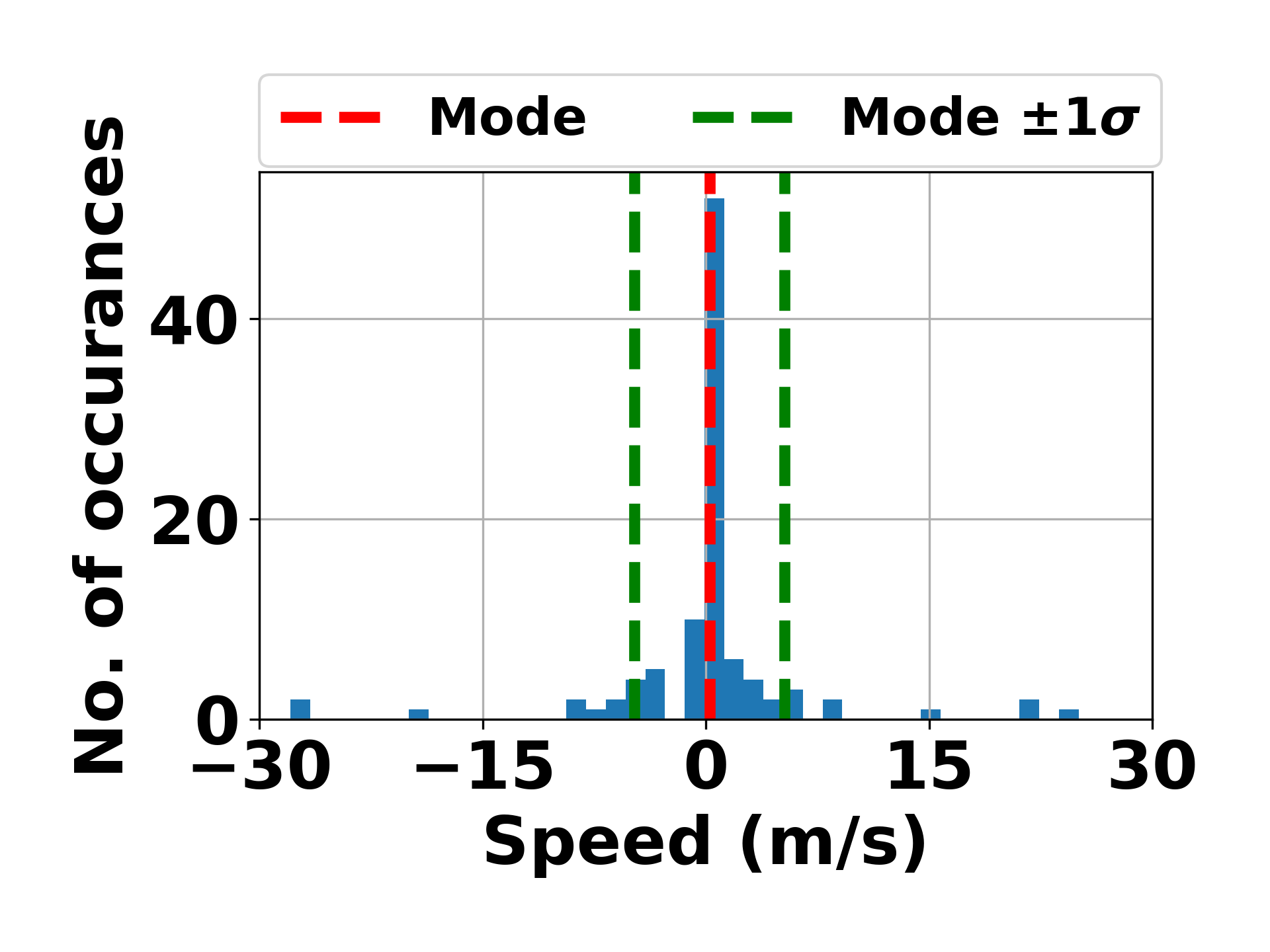}\label{fig:vel_hist}
    }
    \hfill
    \subfigure[]{
    \includegraphics[width=0.43\textwidth]{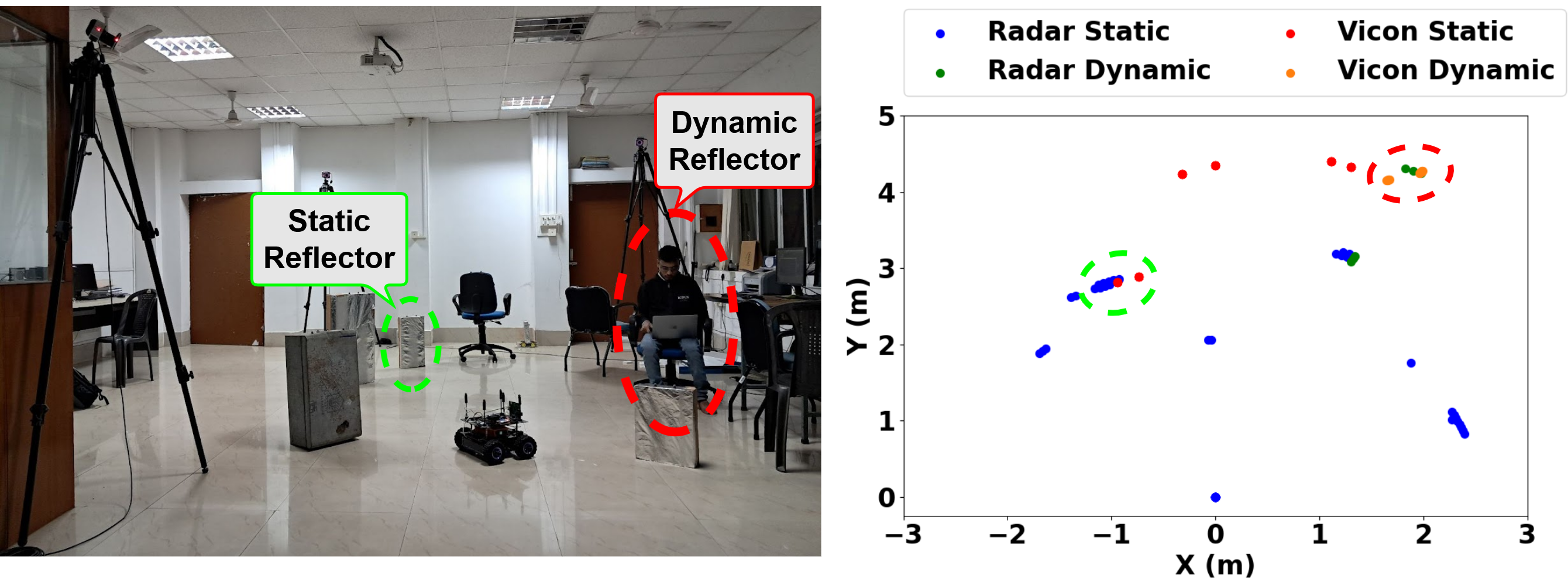}
    \label{fig:static_dynamic}}
    \caption{(a) For all static points, the radial component (green arrow) of the relative speed which varies sinusoidally, (b) Distribution of ego-vehicle speed estimates, (c) A typical dynamic scene (left) and classification output (right).}
    \label{fig:static_dynamic_method}
    \vspace{-0.5cm}
\end{figure*}

\subsection{Design Choices}

Traditional radar applications often depend on large, costly scanning radars that rotate and scan their environment using mechanical motors. We exclude these radars from our study as they are unsuitable for lightweight, compact robots. Instead, we focus on single-chip COTS radars. Unlike previous approaches that rely on additional sensors such as wheel encoders, IMUs, or LiDARs~\cite{lu2020milliego,lu2020see,kramer2020radar,sen2024continuous, almalioglu2020milli,huang2024less}, \ourmethod leverages the intrinsic properties of radar for odometry. This choice is justified as IMU-based speed estimation leads to integration error accumulation, and radar's inherent capabilities make IMUs redundant for 2D SLAM. Most existing ego-motion estimation methods are designed for environments with static objects within the FoV~\cite{kramer2020radar,sie2024radarize}. However, in real-world scenarios, dynamic objects are commonly present. Our design addresses this challenge by incorporating dynamic object handling into our ego-speed estimation framework. Most previous works rely on mmWave radar data combined with DNNs to estimate translational speed~\cite{sie2024radarize,sie2023batmobility,lu2020milliego}. However, these methods are computationally intensive and challenging to implement on edge devices for real-time computing. Our approach demonstrates how signal processing and mathematical approaches can be effectively used to achieve precise ego-speed estimation.

\subsection{Design Outline}
As shown in \figurename~\ref{fig:pipeline}, our method is designed with two main components that work together to estimate the speed of an ego-vehicle using radar data. Each component addresses a specific aspect of motion estimation, leveraging different properties of radar sensing. The components are as follows:

\noindent\textbf{Differentiating between static and dynamic objects:} The initial step involves distinguishing between static and dynamic objects within the FoV by analyzing the radial speed profiles of individual objects. When a radar sensor moves, all stationary targets appear to move in the opposite direction from the sensor's perspective. To effectively separate static from dynamic objects, we employ distribution-based thresholding on the scaled radial velocities of all points in the radar PCD. Since static points have a similar distribution in the radial speed profiles relative to the radar thus, it effectively segregates the static objects from dynamic ones.

\noindent\textbf{Translational Speed Computation:} We use a novel phase-based approach to compute the translational speed of the ego-vehicle. Unlike previous methods that rely on doppler shift~\cite{sie2024radarize,sie2023batmobility}, which may not be effective at low speeds (sub-doppler resolution) and can be erroneous due to PCD estimation, our approach utilizes the subtle phase changes in reflectors within the environment. These phase variations provide precise relative movement information of the vehicle, allowing accurate speed estimation in slow-speed applications.

\subsection{Segregating Static Objects from Dynamic Objects}
We differentiate between static and dynamic objects in the radar's FoV by analyzing the relative speed profiles of individual points detected by the radar. For a stationary radar, the doppler shift directly gives the radial speed of moving objects. But when the radar is in motion, all targets within its FoV appear to move relative to the radar. For stationary objects, the doppler shifts follow a cosine relation with the angle between the radar and the object. However, for dynamic objects, the radial speed has components from both the radar velocity and the object's own velocity and, therefore, does not follow the direct cosine relation. We exploit this property to segregate static and dynamic objects.  

The radar return for a frame can be expressed as $\mathcal{P} = \{\mathbf{p}_i = (r_i, \theta_i, \phi_i, {v_r}_i)\}_{i=1}^{N}$, where the $i^{th}$ point in the radar PCD is characterized by its range $r_i$, azimuth angle $\theta_i$, and elevation angle $\phi_i$, in the radar's frame of reference, and ${v_r}_i$ is the radial speed of the point, relative to the radar. Consider \figurename~\ref{fig:radial}, where a radar-mounted ego-vehicle is moving in an environment with a static ($P_1$) and a dynamic ($P_2$) object in the FoV of the radar. The instantaneous velocity of the ego-vehicle is $v_b$, making an angle $\alpha$ with the radar pointing direction. The dynamic object, $P_2$, moves with a velocity $v_d$, making an angle $\gamma$ with the radar pointing direction. The radial speeds of the static and dynamic points can be given by ${v_r}_1 = v_b cos(\theta_1 - \alpha)$ and ${v_r}_2 = v_b cos(\theta_2 + \alpha) - v_d cos(\theta_2 + \gamma)$, respectively. Notably, all static points will have similar radial velocity profiles ${v_r}_i = v_b cos(\theta_i - \alpha)$. However, dynamic points will have different velocity profiles due to the contribution from their velocity component, ${v_d}_i cos(\theta_i + \gamma_i)$. We define the ego-vehicle speed estimate from the $i^{th}$ point, $\hat{v_b}_i$ as $\frac{{v_r}_i}{cos(\theta_i - \alpha)}$. We observe that for static objects, $\hat{v_b}_i$ will be theoretically equal to ego-vehicle speed, $v_b$, whereas dynamic objects will not follow this relationship. In practice, $\hat{v_b}_i$'s do not exactly equal $v_b$ due to the radar's noise and doppler and angular resolutions. The static points are identified as,

\vspace{-0.25cm}
\begin{equation}
    S = \{ \mathbf{p}_i \in \mathcal{P} \; | \; \hat{v_b}_i \in (M_0 - k\sigma, M_0 + k\sigma)\}
\end{equation}

where $M_0 = \operatorname{mode} \limits_{i} (\hat{v_b}_i)$, $\sigma$ is the standard deviation, and $k$ is a multiplying factor which signifies the spread of the static points. Higher errors in doppler and angular estimation lead to higher values of $k$. We have chosen $k$ as $1$. Figure \ref{fig:vel_hist} shows the distribution of ego-velocity estimates for a frame. Estimates corresponding to dynamic points lie away from the peak corresponding to static points. We assume static points outnumber dynamic points, which is fair since indoor scenarios contain static reflectors like walls, windows, and furniture with high radar cross-sections.

To find $\alpha$, consider two static points $\mathbf{p}_i$ and $\mathbf{p}_j$. The ratio of their radial velocities can be expressed as $\frac{{v_r}_i}{{v_r}_j} = \frac{cos( \theta_i - \alpha )}{ cos( \theta_j - \alpha) }$. We obtain an estimate of $\alpha$ as,
    
\begin{equation}\label{eq:finding_alpha}
    \hat{\alpha}_{i,j} = \frac{ {v_r}_j cos(\theta_i) - {v_r}_i cos(\theta_j) }{ {v_r}_i sin(\theta_j) - {v_r}_j sin(\theta_i) }
\end{equation}

With the assumption that static points are in the majority, we can estimate $\alpha$ as $\hat{\alpha} = \operatorname{mode} \limits_{\substack{i, j \\ i \neq j}} (\hat{\alpha}_{i,j})$.


\subsection{Estimating Translational Changes with Phase Unwrapping} 
Once we have the static points, we estimate the translational speed of the ego-vehicle. For that purpose, we first need to precisely track the presence of the objects in the FoV. 
\subsubsection{Range bin selection}

The range-FFT on the radar's raw I/Q frames outputs the positions and phase variations of the objects present in the FoV of the radar. Initially, we identify the peak index in the range-FFT for a given frame and chirp, which corresponds to the object's distance from the radar based on the $95^{th}$ percentile of the signal-to-noise (SNR) value. As identified in the previous stage, we select those peak indices closer (within $\pm$ 3 range bins) or exactly at the range where static objects are present. 
Once we select the range bins corresponding to the static objects, our next objective is to analyze the phase variation in those bins over time.

\begin{figure}[!t]
    \centering
    \subfigure[]{
    \includegraphics[width=0.25\textwidth]{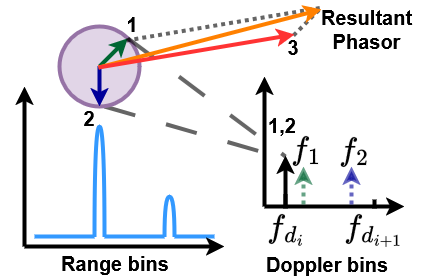}\label{fig:phasor}
    }\hfill
    \subfigure[]{
    \includegraphics[width=0.21\textwidth]{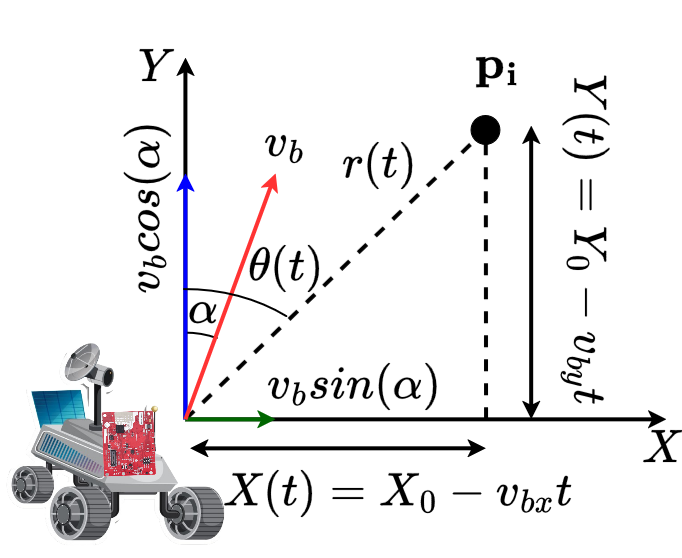}
    \label{fig:scene}}
    \caption{(a) Speeds falling between two doppler bins are assigned to the lower bin due to FFT; resultant of two phasors is shown, (b) A typical scenario where a static object is present in front of a moving radar.}
    \label{fig:standard_methd}
    \vspace{-0.5cm}
\end{figure}

\subsubsection{Phase Unwrapping of Raw Phase Values}

The raw phase values obtained from the radar at a particular range bin where a static object is present may experience discontinuities due to the periodic nature of the phase representation. To ensure accurate phase measurements for speed estimation, we perform phase unwrapping. This process involves identifying and correcting phase jumps greater than $\pi$ radians, providing a continuous phase representation over time. Specifically, if a phase difference between consecutive measurements exceeds $\pi$, a correction is applied by subtracting multiples of $2\pi$ to eliminate the discontinuity. The unwrapped phase can be expressed as: $\Phi_{\text{unwrapped}}(t) = \Phi(t) + 2\pi n$, where $n$ is an integer that adjusts the phase to remove discontinuities. Due to phase unwrapping, we can only estimate velocities without ambiguity when the consecutive change in the phase values, $\Delta \Phi$, is less than $2\pi$. Thus, unwrapping the phase values leads to a maximum measurable speed of up to $7.75$ m/s.

\subsubsection{Computing Translational Speed}
As shown in \figurename\ref{fig:radial}, the previous works on ego-speed estimation use the doppler shifts from the radar PCD and compute the ego-speed as $v_b = \frac{{v_r}^i}{cos\theta^i}$, where ${v_r}^i$ and $\theta^i$ are the radial speed and angle-of-arrival of $i^{th}$ PCD respectively. As PCD estimation is sparse due to poor angular resolution $(\approx15 \degree)$, it can easily corrupt the final translational speed computation. On the other hand, the doppler shift ${v_r}^i$ is computed by doppler-FFT on the number of chirp axis of the radar raw I/Q data, so if the phase change in the chirp axis is lower than the doppler resolution then FFT-based doppler speed computation won't detect the speed accurately. Thus, both these approaches together contribute to an erroneous ego-speed estimation. 

Our idea for ego-speed estimation is derived from the basic physics of the phase values. As shown in \figurename~\ref{fig:scene}, at time $t$ within a frame $(0 < t < T)$, we have the static object at a distance of $r(t)$, given by $r(t)^2 = X(t)^2 + Y(t)^2$. As seen previously, the ego-speed vector has magnitude $v_b$ and makes an angle $\alpha$ with the radar pointing direction. 
On differentiating with respect to time $(t)$ and assuming ego-speed, $v_b$ remains constant during the frame time, $T_f$ (which is $150$ ms in our case) we have,
\begin{equation}\label{eq:2}
\begin{aligned}
r(t)\frac{dr(t)}{dt} = (X_0 - v_b cos(\alpha) t)(-v_b cos(\alpha)) \\ + (Y_0 - v_b sin(\alpha) t)(-v_b sin(\alpha))
\end{aligned}
\end{equation}
On integrating Eq.~\eqref{eq:2} with respect to time from $t=0$ to $t$ and with respect to $r(t)$ from $r_{0}$ to $r_t$ we have,

\begin{equation}\label{eq:3}
    r_t = \sqrt{ r_0^2 - 2 v_b t K(\alpha) - v_b^2 t^2}
\end{equation}


where $K(\alpha) = Y_0 cos(\alpha) + X_0 sin(\alpha)$. When the ego-vehicle is in motion, the distance $r(t)$ changes with time, leading to a corresponding phase change  $\Phi(t) = \Phi_0 + \frac{4 \pi r}{\lambda}$. Therefore, the rate of change of $r_t$ at multiple chirps coming within a single frame can be expressed as $\frac{dr_t}{dt} = c \frac{d\Phi(t)}{dt}$, where  $c$ is a constant. After differentiating both sides of Eq.~\eqref{eq:3} with respect to $t$, squaring and simplifying, we have,
\begin{equation}\label{eq:final}
\begin{split}
    a(t) \: v_b^4 + b(t, \alpha) \: v_b^3+  c(t, \alpha) \: v_b^2 + d(t, \alpha) \: v_b + e(t) = 0 
\end{split}
\end{equation}
where $a(.), b(.), c(.), d(.)$ and $e(.)$ are coefficients given by, 

\vspace{-0.3cm}
\begin{align*}
    \begin{array}{l}
        a(t) = t^2, \quad \quad \quad \quad \quad \quad \quad \; \;
        b(t, \alpha) = -2K(\alpha)t \\[5pt]
        c(t, \alpha) = K(\alpha)^2 - t^2c^2\Theta^2(t) \\[5pt]
        d(t, \alpha) = 2tK(\alpha)c^2\Theta^2(t), \quad
        e(t) = -r_0^2 c^2 \Theta^2(t)
    \end{array}
\end{align*}

where, $\Theta(t) = \frac{d\Phi}{dt}(t)$. Eq.~\eqref{eq:final} represents a 4th-order equation in ego-speed, $v_b$, with four possible roots. For each static point, $j$ from $1$ to $N$, we directly obtain $\Phi_{i, j}$ at a time instant $t_i$, where $i$ goes from $1$ to $N_c$ (number of transmitted chirps). For each static point in a frame, we have four roots of the equation for each chirp. Thus, we have $N \times N_c \times 4$ such roots in a single frame. One root must be common across all the chirps, as $v_b$ is a physical quantity. We compute all the roots for a single frame and take the mode to find the common root, which occurs most frequently. This approach helps compute speed even below doppler resolution as it depends purely on the phase changes of the selected static objects.

Although this approach is effective when a static object is the only one in its range bin. When multiple objects share the same range bin, their phasor representation forms a resultant phasor that deviates from Eq.~\eqref{eq:3}, imposing a restrictive condition on selecting static points. This constraint can be relaxed by noting that the resultant of two phasors, with amplitudes $A_1 >> A_2$, closely follows the dominant phasor, as shown in \figurename\ref{fig:phasor} by phasors $1$ and $3$. For example, let's say we have $n$ objects in a single range bin, whose phasors can be computed with doppler-FFT having amplitudes $A_1, A_2, ..., A_n$. We consider this range bin for further computation only if a single phasor is dominant over the other phasors ($A_1 >> A_2 >> ... >> A_n$). The advantage of our method over naive doppler-based processing is demonstrated using \figurename\ref{fig:phasor}. For an object located in the indicated range bin, phasor 1 (green) and 2 (blue) show two possible radial speeds ($f_1$ and $f_2$), both of which lie between the $i$ and $(i+1)^{th}$ doppler bins. Doppler-FFT places both speeds in the $i^{th}$ bin, whereas our method can accurately find and distinguish them. 


\section{Implementation}
We have implemented \ourmethod over three different setups, as discussed next. 

\subsection{Hardware setup}
\begin{figure}
    \centering
    \subfigure[]{
    \includegraphics[width=0.23\textwidth]{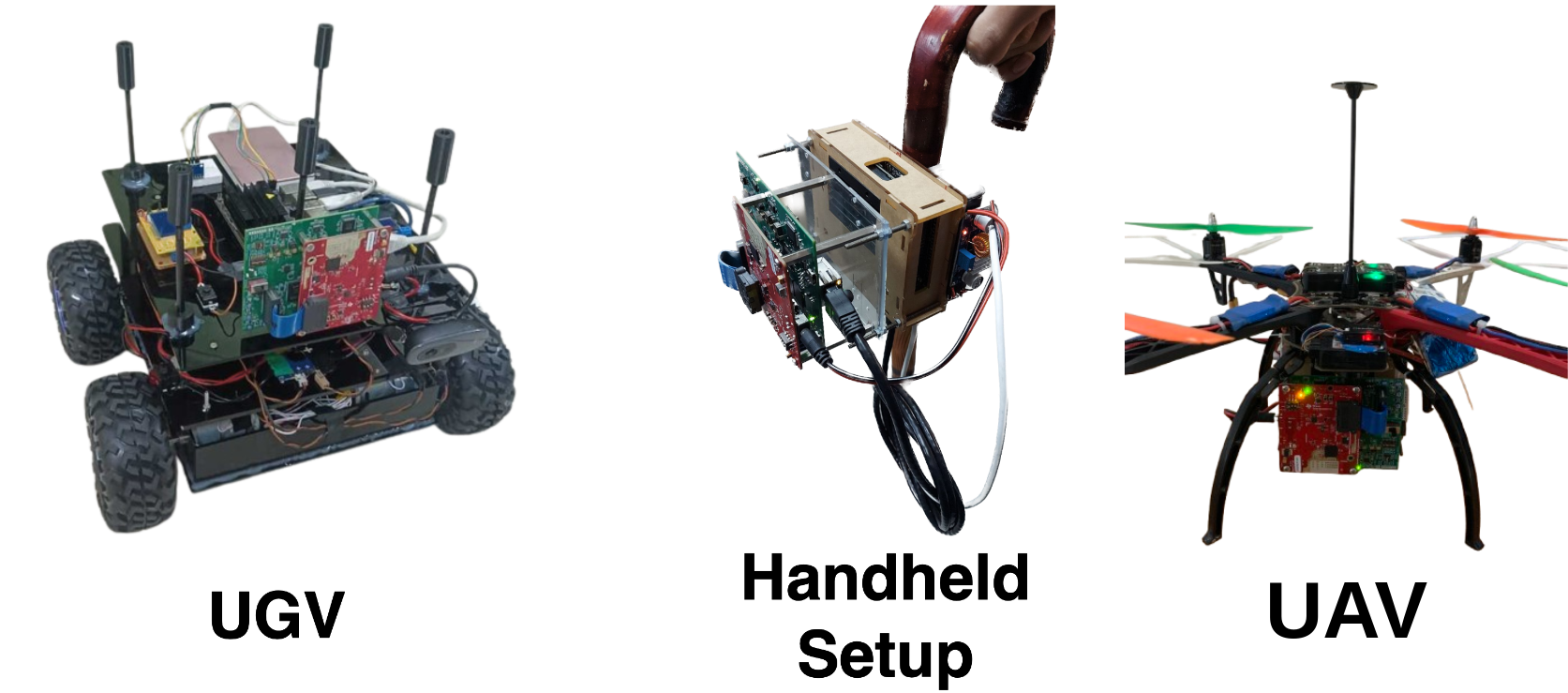}\label{fig:impl}
    }\hfill
    \subfigure[]{
    \includegraphics[width=0.23\textwidth]{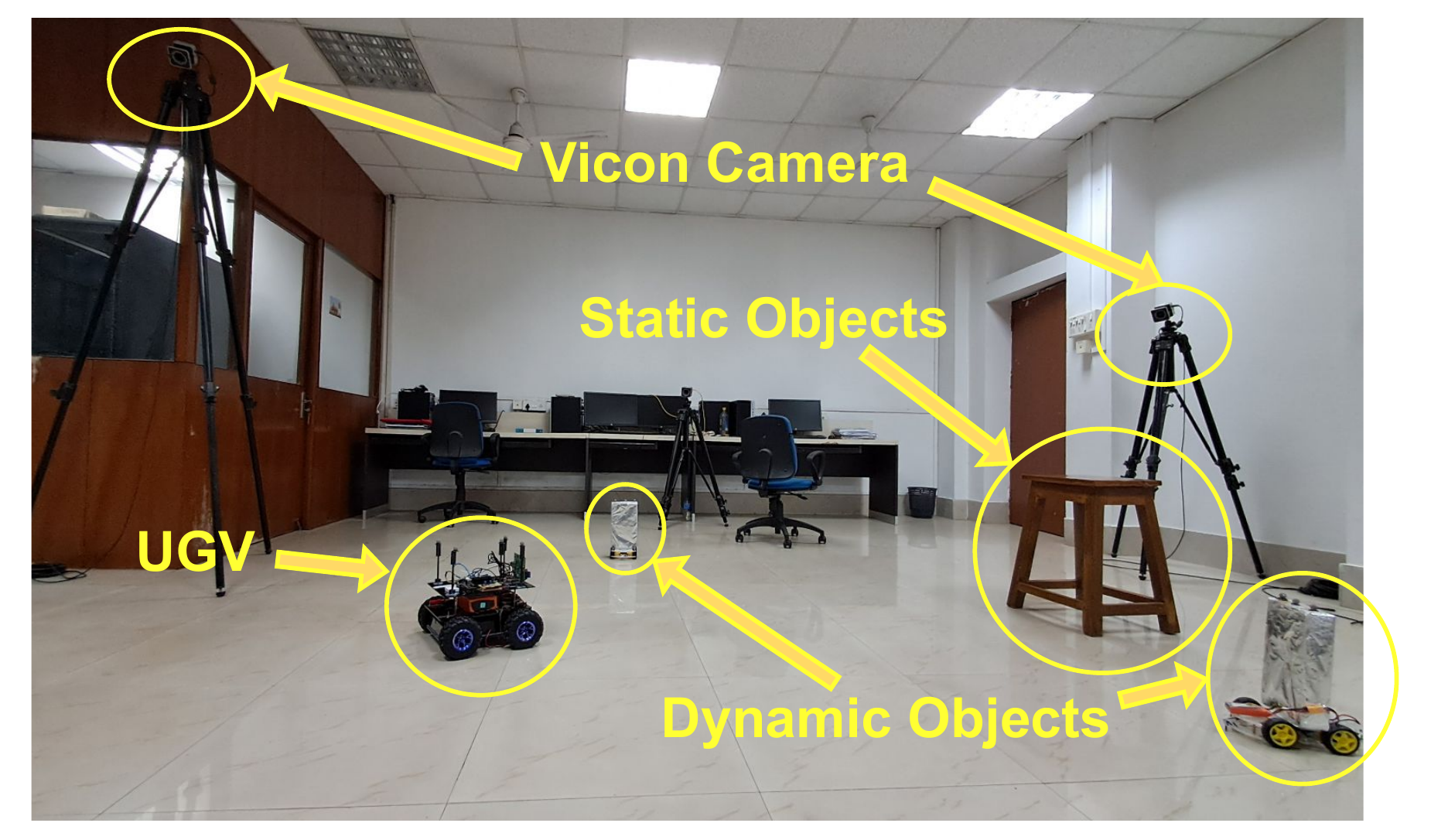}\label{fig:deployment}
    }
    \caption{(a) Different ego-vehicle settings, (b) deployment setup of \ourmethod.}
    \vspace{-0.5cm}
\end{figure}

\noindent
\textbf{Platforms:} We use a Texas Instruments IWR1843 radar~\cite{iwr1843boost}, and a DCA1000
data capture card to receive raw I/Q samples. We evaluate our method on three platforms: (i) \textit{UGV ego-vehicle}: We mount our radar on a remotely operated ground robot, which navigates along various trajectories.  
(ii) \textit{UAV}: We utilize a low-flying drone with a \textit{Pixhawk 2.4.8 flight controller} to collect radar data under different speeds. (iii) \textit{Handheld}: For the handheld setup, we mount the radar on a $95$ cm long handheld stick carried around by volunteers in predefined trajectories. The hardware setups are shown in \figurename~\ref{fig:impl}. In each case, for computing, we use a Jetson Nano running Ubuntu 18.04, featuring a quad-core 
CPU and 4 GB of RAM. 
    We also use a 6-axis IMU, \textit{MPU6050} for logging the acceleration and gyroscope values.




\noindent\textbf{Radar Configuration:} The radar works in a frequency range of 77-81 GHz. We configure the radar to a range resolution of $0.0429$ meters, and the radial speed resolution is set to $0.0496$ m/s. The system operates at $10$ frames per second, with $182$ chirps per frame and $256$ ADC samples per chirp. 

\subsection{Ground Truth Speed Estimation}
To establish the accuracy of our speed measurements, we utilized a Vicon Vero tracker (v1.3X) motion capture system, which provides high-precision, real-time tracking of the position and orientation of the ego-vehicle. By comparing the speed estimates from the radar with the ground truth measurements from the Vicon system, we could evaluate the accuracy and reliability of our speed estimation algorithm.

\subsection{Baselines and Evaluation Metrics}
We choose the following baselines for comparison of our method: (1) Radarize~\cite{sie2024radarize}, which uses a ResNet18-based DNN architecture for translational speed estimation taking only radar doppler-azimuth heatmaps as its input, (ii) Doppler based approach~\cite{rao2017introduction}, which computes doppler-FFT on the raw I/Q data and estimates the speed from the peak doppler bins, (ii) IMU-based odometry ~\cite[Sec. 6.1.1]{lei2018imu} and (iii) MilliEgo~\cite{lu2020milliego} model which fuses both
mmWave range-angle heatmaps along with IMU. It predicts the position coordinates of the ego-vehicle, and we modify it to predict the ego-speed instead. We use the Vicon system to capture the trajectory of our ego-vehicle. From this recorded trajectory, we calculate the ground truth speed of the ego-vehicle. We have considered mean absolute error as the primary evaluation metric.
\section{Evaluation}
This section provides a thorough evaluation under different combinations of reflectors and ego-vehicle setups. 




\subsection{Overall Performance}
\begin{figure}[!t]
    \centering
    \includegraphics[trim=25 15 25 25, clip, width=0.25\textwidth]{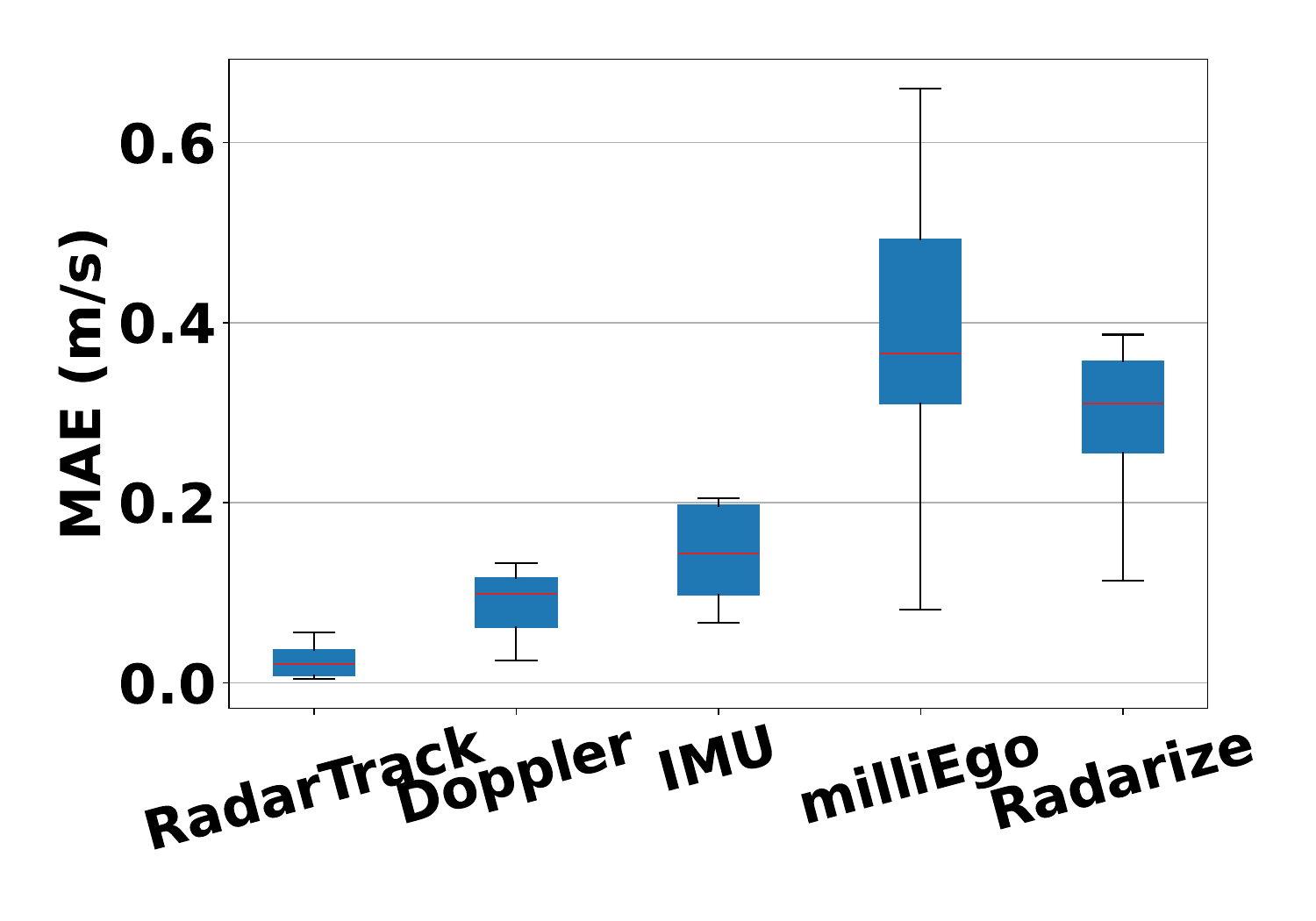}
    \caption{MAE with \ourmethod compared to baselines.}
    \label{fig:box_overall}
    \hspace{-0.1cm}
\end{figure}

\begin{figure}[!t]
    \centering
    \subfigure[]{
    \includegraphics[trim=25 15 25 20, clip,width=0.22\textwidth]{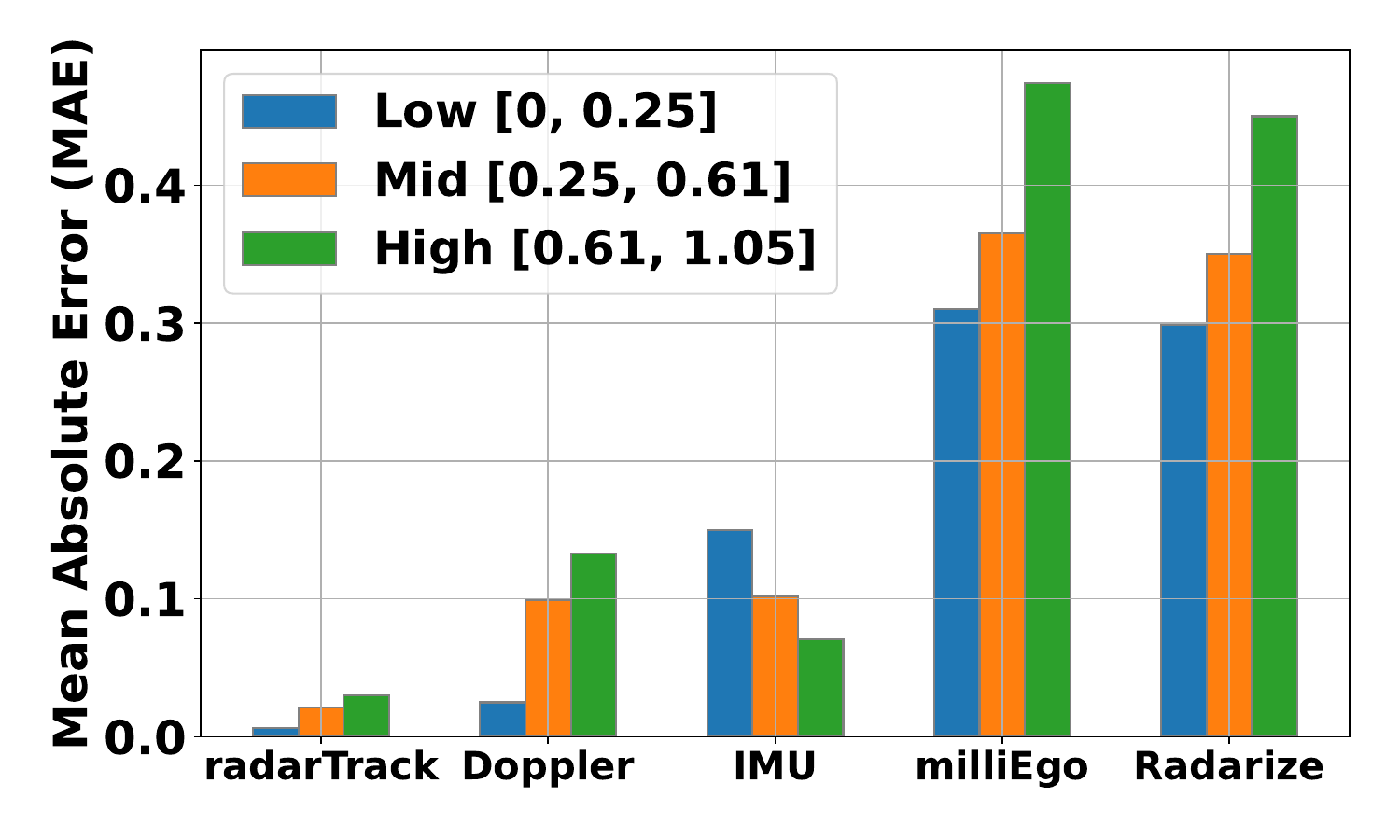}\label{fig:res_across}
    }\hfil
    \subfigure[]{
    \includegraphics[trim=25 15 25 25, clip,width=0.23\textwidth]{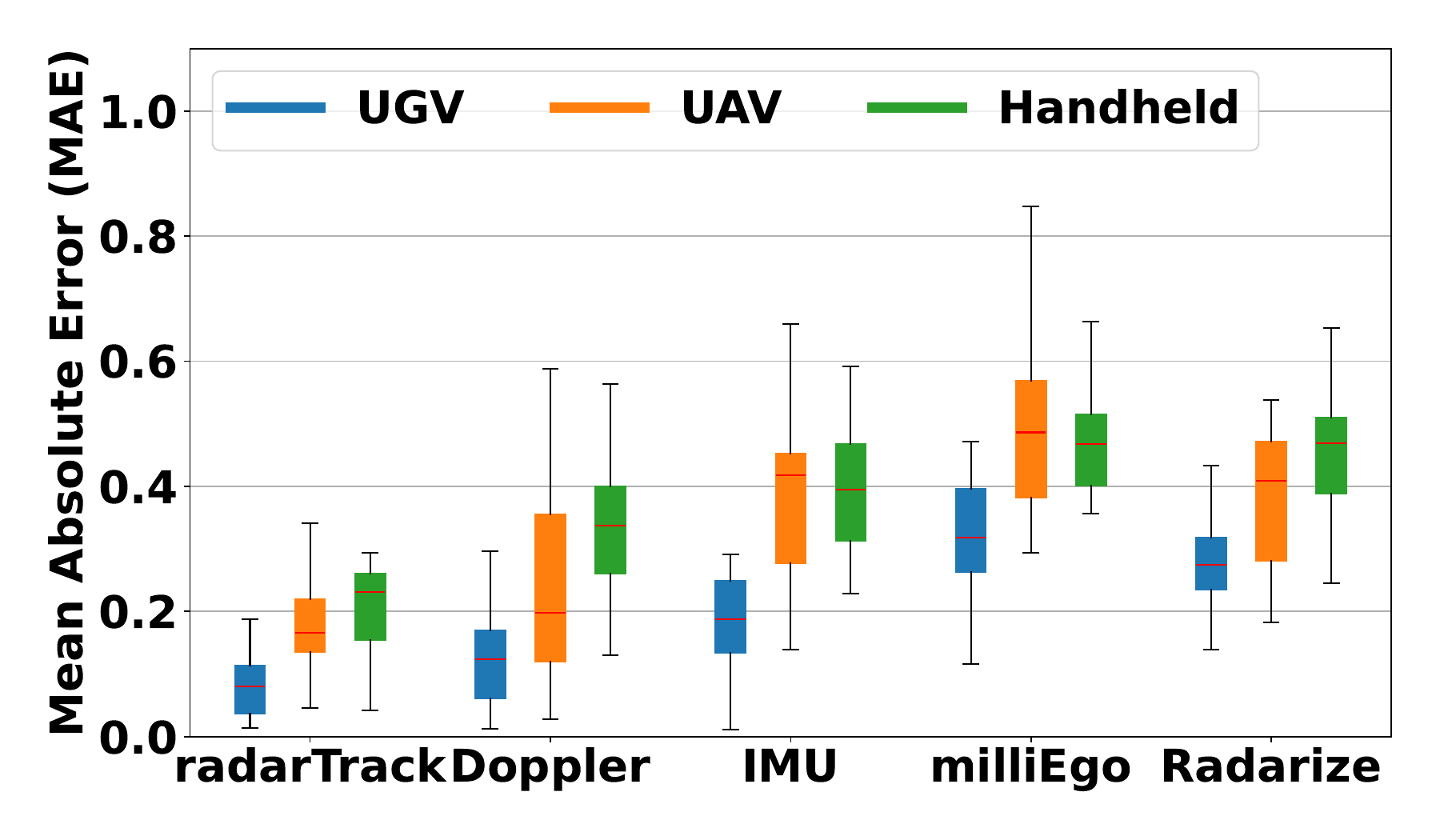}\label{fig:egovehicle}
    }
    \caption{MAE across (a) different speeds, (b) different ego-vehicle platforms.}
    \label{fig:result}
    \vspace{-0.5cm}
\end{figure}

\figurename~\ref{fig:box_overall} shows the distribution of MAE across different methods. As can be seen, the median of MAE with \ourmethod is $\approx 2$ cm/s. \figurename~\ref{fig:res_across} presents the MAE for various methods at different speed levels (Low, Mid, High). The speed ranges are defined as follows: \textbf{Low} $<$ 0.25 m/s, 0.25 m/s $<$ \textbf{Mid} $<$ 0.61 m/s, and 0.61 m/s < \textbf{High} $<$ 1.05 m/s. Our method consistently achieves the lowest error (approx $5\%$ of the base speed, on average), outperforming the closest competitor, the doppler-based approach, by up to 4x in accuracy.

As the vehicle's speed increases, we notice that the error grows, likely due to the increasing noise in the phase component of the reflected signals. The doppler-FFT technique also tends to worsen speed estimation at higher speeds as the inherent phase errors can accumulate more errors in the doppler-FFT. Also, at low speeds, the doppler-based method struggles to capture small-scale movements that fall below the doppler resolution, leading to higher errors when compared to our approach. Interestingly, the IMU-based odometry performs better at high speeds compared to low speeds. This discrepancy is likely caused by time drift in the IMU data during lower-speed scenarios, where longer collection durations introduce drift errors. milliEgo~\cite{lu2020milliego} performs the worst, suffering from limitations in both IMU and mmWave modalities. On the other hand, radarize~\cite{sie2024radarize} also performs poorly during dynamic scenarios when multiple objects are moving in the radar's FoV. 

\subsection{Performance with Different Ego-Vehicle Settings} 

We evaluated the performance of \ourmethod on three different ego-vehicle platforms: UGV, handheld device, and UAV, as demonstrated in \figurename~\ref{fig:impl}. \figurename~\ref{fig:egovehicle} presents the results for \ourmethod across the different ego-vehicle platforms, comparing its performance against other baselines. For the UGV platform with a baseline speed of $0.25$ m/s, \ourmethod achieves a MAE of approximately $0.02$ m/s, performing consistently well due to the minimal noise caused by translational movement on the ground. In contrast, the UAV platform (baseline speed of $0.21$ m/s) presents more challenges, as vibrational noise and translational motion in the vertical axis introduce significant noise into the speed estimates. Consequently, the MAE for the UAV setup is around $0.15$ m/s, which is notably higher due to the difficulty in compensating for the UAV's variable altitude and dynamic movement. Lastly, the handheld setup (baseline speed of $0.35$ m/s) exhibits an intermediate level of noise, primarily due to human subjects repeatedly turning the stick left and right, generating noise in the speed estimates, leading to an MAE of approximately $0.16$ m/s. 


\subsection{Performance Improvement with Phase-Based Approach at Different Speeds}

State-of-the-art methods for ego-speed estimation, such as Radarize~\cite{sie2024radarize} and MilliEgo~\cite{lu2020milliego}, primarily rely on range-doppler processing, which involves applying an FFT to the range data across multiple chirps. However, this approach results in a fixed doppler resolution, meaning speeds that fall between the resolution steps cannot be accurately estimated.

In contrast, using raw phase data enables tracking much finer movements due to the relationship $\phi = \frac{4\pi d}{\lambda}$, where $\lambda$ is the wavelength, typically on the order of millimeters ($\approx$ 4 mm). So by leveraging the phase variations, for example with a phase change of just $0.057^\circ$, minute changes in range (as small as $\Delta r \approx 0.63 \mu m$) can be detected~\cite[Sec. II-B]{basak2022mmspy}. 

\begin{figure}[!t]
    \centering
    \subfigure[]{
    \includegraphics[trim=25 15 25 25, clip,width=0.23\textwidth]{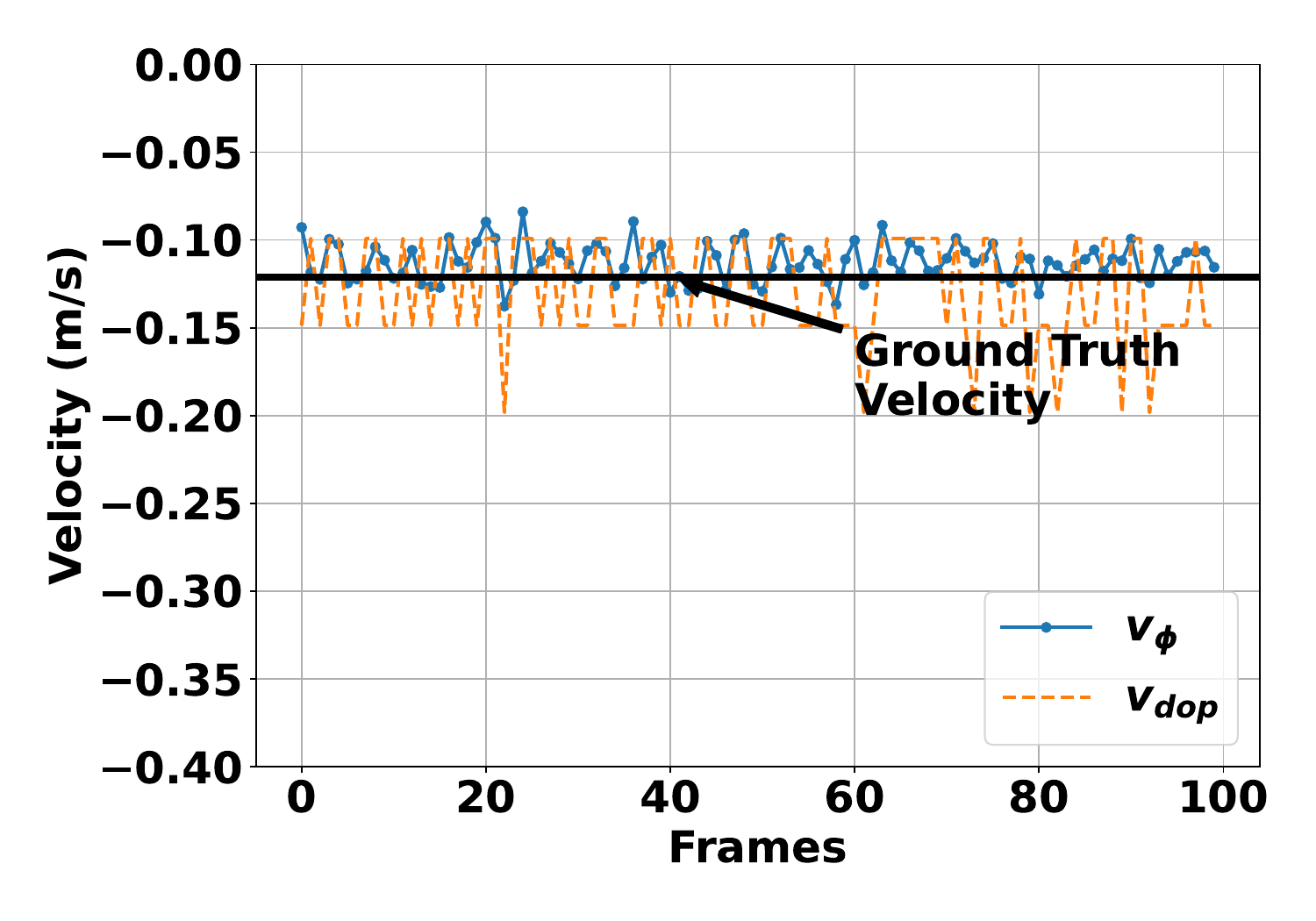}
    }\hfil
    \subfigure[]{
    \includegraphics[trim=25 15 25 25, clip,width=0.23\textwidth]{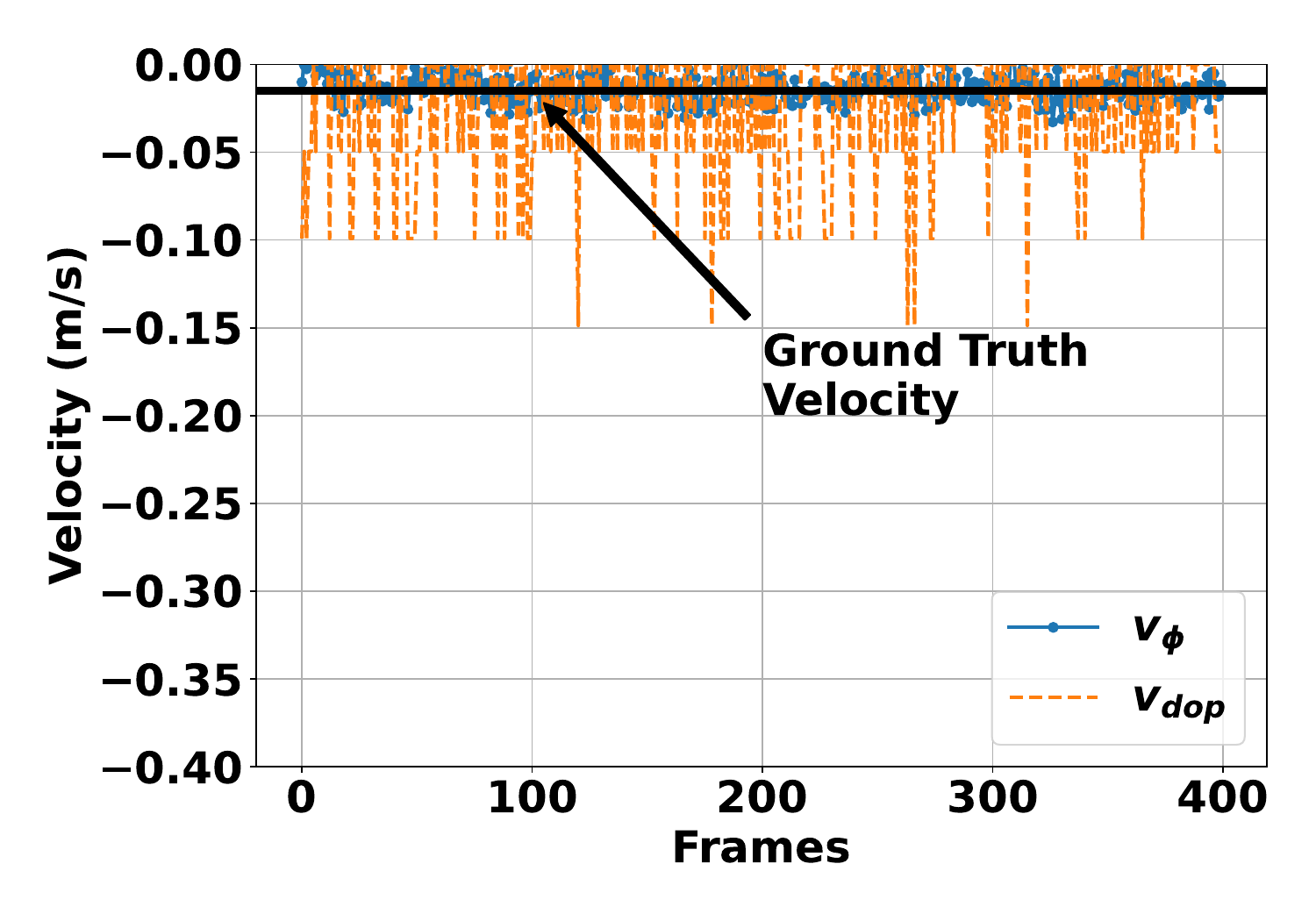}
    }
    \caption{Comparative study of the doppler-based vs phase-based speed computation: (a) Ego-vehicle moving at medium speed (12 cm/s) and (b) lower speed (1.5 cm/s) (Note: $^-$ve speed represents the ego-vehicle is approaching a target).}
    \label{fig:phase_vs_dop}
    \vspace{-0.5cm}
\end{figure}

In \figurename~\ref{fig:phase_vs_dop}, we present a qualitative comparison between speed estimates from the phase-based method and those obtained from the range-doppler heatmap, a conventional state-of-the-art approach. The speed estimates from the range-doppler heatmap are notably noisier at lower speeds (e.g., 1.5 cm/s) because the true speed lies within the doppler resolution limit (4 cm/s). Even at higher speeds, the phase-based method delivers more accurate speed estimates. This is because, at high speeds, the range-doppler FFT can introduce errors as the range bin also shifts within a single frame, leading to incorrect phase values when tracked at the frame level. In contrast, \ourmethod operates at a chirp-level resolution for phase-based speed estimation, enabling more accurate and robust performance across a range of speeds.

\subsection{Performance with different combinations of static and dynamic reflectors}


We evaluate the performance of \ourmethod under various combinations of static and dynamic reflectors (see \figurename~\ref{fig:deployment}): (i) 3 static objects, (ii) 2 static and 1 dynamic object, (iii) 2 static and 2 dynamic objects, (iv) 1 static and 1 dynamic object, and (v) 2 dynamic objects.

\begin{table}[htbp] 
\centering 
\caption{MAE for Different Combinations of Static and Dynamic Reflectors} 
\label{tab:reflector} 
\begin{tabular}{|l|l|} 
\hline \textbf{Configuration} & \textbf{MAE (m/s)} \\ \hline 
3 static objects & \textbf{0.018} \\ \hline 2 static objects, 1 dynamic object & 0.025 \\ \hline
2 static objects, 2 dynamic objects & 0.037 \\ \hline 
1 static object, 1 dynamic object & 0.045 \\ \hline
2 dynamic objects & 0.054 \\ \hline 
\end{tabular} 
\end{table}

The static objects in our setup include typical household items like chairs and metal objects, representing surfaces with different reflective properties. For dynamic objects, we utilize a small robot car with an aluminum foil-coated box (to emulate high reflectivity to generate maximal interference with our method) and human subjects to capture different motion and reflective behaviors. The accuracy of \ourmethod tends to decrease as the number of high-reflective dynamic objects increases. This is primarily due to the additional phase noises introduced by the movement of dynamic objects, which complicates the inference process. However, despite the increasing complexity, \ourmethod consistently maintained a reasonable level of performance, effectively handling the dynamic environments. The results (see Table~\ref{tab:reflector}) highlight that introducing dynamic objects leads to some degradation in accuracy, but the overall system remains robust. Notably, with three static objects, the system achieves its highest accuracy since the reflections from the static objects remain stable. Conversely, with two dynamic objects, performance dips slightly due to the increased unpredictability and motion interference in the radar FoV. Notably, the ego-vehicle does not observe any static objects initially in this case, thus having a high estimation error initially (approx $0.2$ m/s); however, as it moves towards the wall, it considers the wall as a static object and reduces the error significantly (approx $0.01$ m/s).

\subsection{Resource and Power Consumption}
\begin{figure}[!t]
    \centering
    \subfigure[]{
    \includegraphics[trim=25 15 25 25, clip,width=0.26\textwidth]{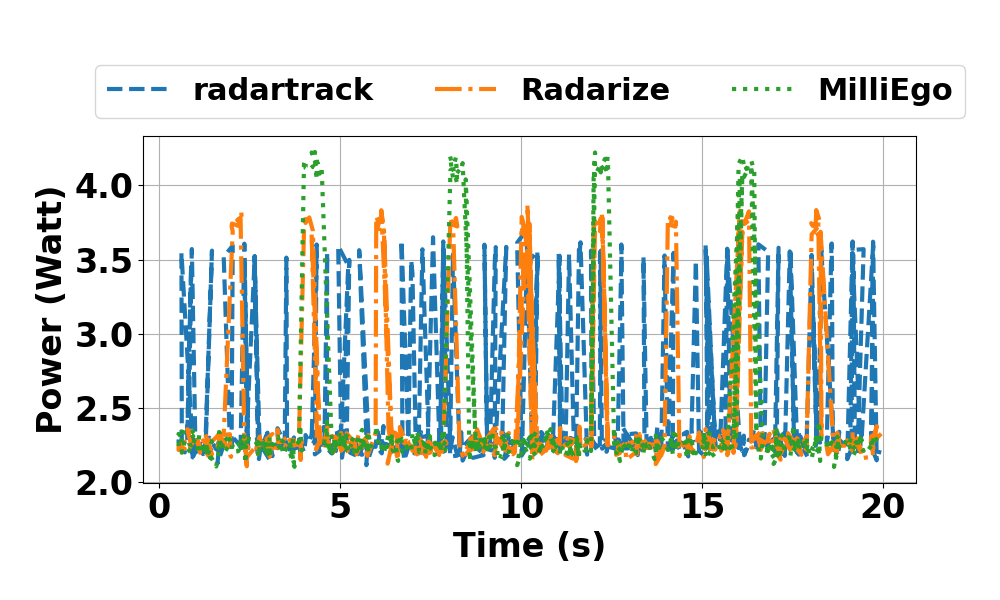}\label{fig:power_latency}
    }\hfil
    \subfigure[]{
    \includegraphics[trim=25 15 25 25, clip,width=0.20\textwidth]{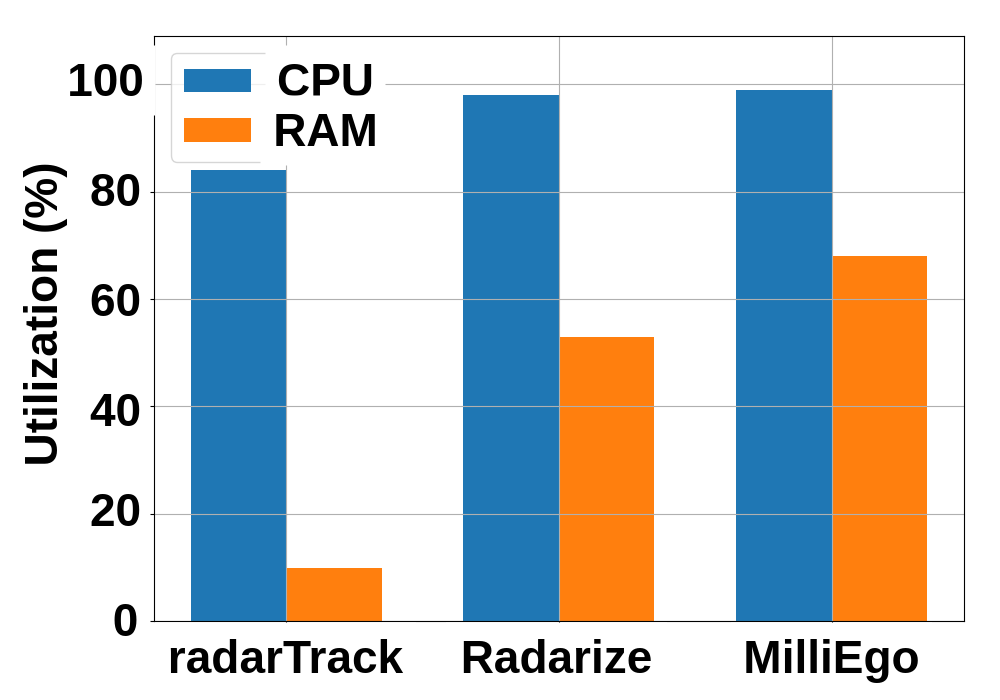}\label{fig:cpu_ram}
    }
    \caption{Power \& computing resource consumption at the edge.}
    \label{fig:pow_reso}
    \vspace{-0.5cm}
\end{figure}

We analyze the power and resource consumption of \ourmethod~in comparison to Radarize and MilliEgo, evaluated on a Jetson Nano equipped with 4 GB RAM and a quad-core CPU. As shown in \figurename~\ref{fig:power_latency}, \ourmethod demonstrates a significantly lower power consumption profile. \ourmethod consumes around 2.25 W power, with occasional spikes reaching 2.56 W during inference. In contrast, the DNN-based methods, Radarize and MilliEgo, exhibit higher peaks, reaching up to 2.75 W and 3.12 W, respectively, making them less suitable for lightweight, low-power applications. As shown in \figurename~\ref{fig:cpu_ram}, \ourmethod utilizes only 84\% of CPU and 10\% of RAM, significantly lower than the DNN-based approaches consuming 99\% of CPU usage, with Radarize using 53\% and MilliEgo consuming up to 68\% of the available 4 GB memory. The latency of \ourmethod is approximately $0.29$ s, significantly lower than that of Radarize ($3$ s) and MilliEgo ($5$ s).

\section{Conclusion}
In conclusion, we propose \ourmethod{}, a novel odometry framework that leverages a single-chip mmWave radar for robust ego-motion estimation in challenging environments. Unlike existing methods, \ourmethod{} relies solely on mmWave radar data, overcoming limitations such as noise, sparsity, and slow speed operation. Our approach, based on phase-based speed estimation, offers low latency and high reliability, making it suitable for mobile and resource-constrained devices. The proposed approach can be extended for real-time navigation support for micro-robots, autonomous maneuvers, and assistive movements through augmented reality, which we keep as the future extension of this work.  


\section*{Acknowledgment}
 This work has been supported by the Department of Science and Technology (NGP Division) with funding approval number NGP/GS-02/Sandip/IIT-K/WB/2023 (C), dated 09-01-2024. The works of Soham Chakraborty and Soham Tripathy are also supported by CHANAKYA Fellowship Program 2023 from the TIH Foundation for IoT \& IoE, IIT Bombay, India.  

\balance
\bibliographystyle{IEEEtran}
\bibliography{main}

\end{document}